\documentclass[twoside,11pt]{article}

\usepackage{jmlr2e}
\usepackage{amsmath}
\usepackage{caption}
\usepackage{subcaption}
\usepackage{graphicx}
\graphicspath{ {imgs/} }

\firstpageno{1}

\begin{document}

\title{Generalizing Across Multi-Objective Reward Functions in Deep Reinforcement Learning}

\author{}
\author{\name Eli Friedman \email friedm3@cooper.edu \\
      \name Fred Fontaine \email fred@cooper.edu \\
      \addr Department of Electrical Engineering\\
      Cooper Union\\
      41 Cooper Sq, NY 10003, USA}

\maketitle

\begin{abstract}
Many reinforcement-learning researchers treat the reward function as a part of the environment, meaning that the agent can only know the reward of a state if it encounters that state in a trial run.
However, we argue that this is an unnecessary limitation and instead, the reward function should be provided to the learning algorithm.
The advantage is that the algorithm can then use the reward function to check the reward for states that the agent hasn't even encountered yet.
In addition, the algorithm can simultaneously learn policies for multiple reward functions. For each state, the algorithm would calculate the reward using each of the reward functions and add the rewards to its experience replay dataset.
The Hindsight Experience Replay algorithm developed by \citet{HER} does just this, and learns to generalize across a distribution of sparse, goal-based rewards.
We extend this algorithm to linearly-weighted, multi-objective rewards and learn a single policy that can generalize across all linear combinations of the multi-objective reward.
Whereas other multi-objective algorithms teach the Q-function to generalize across the reward weights, our algorithm enables the policy to generalize, and can thus be used with continuous actions.

\end{abstract}


\section{Introduction}
Reinforcement Learning has successfully been used to accomplish many tasks, such as learning to play games \citep{DQN, AlphaGo}, control robots \citep{helicopter, manipulator}, and drive cars \citep{driving}. 

In model-free reinforcement learning, the environment model is considered a black box, so the agent can only interact with it through sample runs. 
The reward function, however, quantifies the task that the problem-designer wants the agent to accomplish, and so is not inherently part of the environment.
Therefore, the reward function could be incorporated into the learning algorithm rather than being included in the environment model.
The algorithm would then be able to check what the reward is for any state or action, even ones it hasn't encountered in its sample runs.
If the designer wants the agent to achieve a high reward for a number of reward functions, the algorithm could check what the reward is for each function and then use an off-policy algorithm to simultaneously learn the value function of each reward.
Alternatively, if the distribution of reward functions can be parameterized, then the parameters can be fed into the policy, which will learn different behaviours for different reward functions.

For example, \citet{HER} develop the Hindsight Experience Replay (HER) method which demonstrates that a neural network policy can generalize across sparse, goal-oriented rewards.
The agent learns how to reach multiple goal states and can even interpolate to goal states that it did not specifically see before.
They add the desired goal to the state representation and augment the experience replay dataset with alternate goals in order to learn to generalize to new goals.

We extend this result to another class of reward functions: multi-objective rewards with a linear scalarization function.
Multi-objective reinforcement learning problems consider environments where the reward is vector valued rather than scalar. \citet{survey} provide a thorough overview.
To solve a multi-objective problem, one can either find all policies on the Pareto front--that is, the set of policies in which no policy is strictly worse than any other--or one can specify a function that converts the reward vector to a scalar. 
The advantage of specifying a scalarization function is that one can then use the standard set of reinforcement learning algorithms to learn how to maximize the reward.
One common scalarization function is a linear function that computes a weighted sum of the elements in the reward vector \citep{linweights1, linweights2, linweights3}.
The disadvantage, however, is that in order to use standard reinforcement learning techniques, the weights must be specified up front, before training the agent.
For example, if you were to design an agent that would help navigate from home to work, you might choose a reward function that weights the commute time, the cost of the route, and the cost to the environment.
Depending on the day, you might consider cost to be more important than time, or vice versa, and would need to retrain the algorithm each time you changed your mind.

We solve this problem by extending Hindsight Experience Replay to the multi-objective reward case.
The weights can then be input to the agent in real time, even after the agent is trained.
Rather than retraining your navigation agent with new priorities, you could just input your new priorities to the agent and it would adjust its policy accordingly.
Our algorithm can be applied to any off-line, deep reinforcement-learning algorithm that uses an experience replay dataset, and can be easily applied with minimal changes to the underlying algorithm.

Ideas similar to ours exist, most notably the work of \citet{MultiFittedQ}, who use fitted Q-iteration \citep{fittedQ} and add the linear weight vector to the state representation in order to learn a Q-function that can generalize across all linear combinations of the reward.
Our solution allows policy functions to generalize across weight vectors, in addition to Q-functions.
Thus, our algorithm can also be applied to environments with continuous action spaces.

\section{Background}
The reinforcement learning problem can be represented as a Markov Decision Process. At each timestep, $t$, the agent observes the environment state $s_t \in S$, chooses an action using its policy $\pi(s_t)$, receives a reward $r_t = r(a_t, s_t)$, and the environment transitions to the next state with probability $T(s_{t+1} | s_t, a_t)$.

The discounted sum of rewards is called the return, $R = \sum_{t=0}^{\infty} \gamma^t r_t $ where $\gamma \in [0, 1)$ is a discounting factor used to ensure that the return converges.
The goal of reinforcement learning is to learn the optimal policy $\pi^*(s_t)$ that maximizes the expected return, $\pi^* = \underset{\pi}{\arg\max} E[R | \pi]$.
The state-action function, or Q function, measures the expected value of a policy from each action from a given state and is defined as $Q_\pi(s_t, a_t) = E[\sum_{i=t}^{\infty} \gamma^{i-t} r_i | s_t, a_t]$

When the environment is specified with a multi-objective reward, $r_t$ becomes a vector $\textbf{r}_t$.
A multi-objective reward will also typically be defined with a scalarization function that maps the vector reward to a single, optimizable scalar.
In this paper, we consider a linear scalar function,
$r_t = \textbf{w}^T \textbf{r}_t$, where $\textbf{w}$ is a set of weights $w_i \in [0, 1]$ such that the weights sum to $1$.

\subsection{Hindsight Experience Replay}

Many neural-network based reinforcement-learning algorithms employ a replay dataset of prior experiences. For example, deep Q-networks \citep{DQN} use Q-learning with a replay dataset and some other tricks to learn an optimal Q function. Actor critic with experience replay \citep{ACER} is an off-policy actor critic method that adds an experience replay dataset and trust-region updates. Deep deterministic policy gradients (DDPG) \citep{DDPG} is an off-policy policy-gradient algorithm for deterministic policies.
The replay dataset contains transition tuples $\langle s_t, a_t, r_t, s_{t+1} \rangle$ that are aggregated into mini-batches and used to fit the neural networks.

\citet{HER} develop HER to deal with the case where the reward is goal-based. The reward is $0$ when the agent reaches the goal state and $-1$ in all other states.
They include the goal, $g$, in the transition tuples--$\langle g, s_t, a_t, r_t, s_{t+1} \rangle$ and feed the goal to the policy and Q function.
When they replay the dataset to the neural network, they swap the original goal state in some of the tuples with the state that the agent actually reached.
This allows the agent to learn both from trial runs where the agent manages to achieve its goal and from the ones where it fails.
In addition, the agent learns to generalize across goals, so that when the agent is online, the user can input a new goal that the agent might never have seen during training.

\section{Reward Generalization}
The data-augmentation technique of Hindsight Experience Replay should not, in principle, be limited to generalizing across goal-based reward functions. Other families of reward functions should also be generalizable, so long as the reward function can be parameterized and fed to the policy and Q functions.

Consider a class of reward function, $r_w = r(s, a, w)$, parameterized by $w$.
Additional rewards can be sampled from this class and added to the replay dataset, so that the agent can learn to generalize across the reward class.
At every timestep, $k$ additional rewards can be chosen from the reward family and add them to the replay dataset $\langle s_t, a_t, w_i, r(s_t, a_t, w_i), s_{t+1} \rangle$ for $i = 1...k$.
The reward parameters $w_i$ are then fed to both the policy network, $\pi(s_t, w_i)$, and the state-action network, $Q(s_t, a_t, w_i)$, and trained as normal.

The agent should be able to generalize across a variety of reward classes. \citet{HER} demonstrate generalization over goal-based rewards. In this work we demonstrate generalization over linearly weighted multi-objective rewards. 
Another reward class that would be interesting to explore would be to generalize over the $\gamma$ parameter, so that the agent could dynamically adjust its reward horizon.

\subsection{Multi-objective Generalization}

We apply this data-augmentation technique to multi-objective rewards with linear weights.
At every timestep, we add $k$ tuples to the replay dataset $\langle s_t, a_t, \textbf{w}_i, \textbf{w}_i ^ T \textbf{r}_t, s_{t+1} \rangle$ for $i = 1...k$.
For example, if $k=2$ and the agent sees state $s_t$, performs action $a_t$, and received reward $\textbf{r}_t$ we can add both the tuple $\langle s_t, a_t, \textbf{w}_1, \textbf{w}_1 ^ T \textbf{r}_t \rangle$, and the tuple $\langle s_t, a_t, \textbf{w}_2\, textbf{w}_2 ^ T \textbf{r}_t \rangle$ to the dataset.
We feed the weight vector to both the policy network, $\pi(s_t, \textbf{w}_t)$, and state-action network, $Q(s_t, \textbf{w}_t, a_t)$, and train as normal using the augmented dataset.

Since the learning algorithm has access to the reward function, it can choose a strategy for sampling from the reward distribution.
For now, though, we uniformly sample the weights we use for augmenting the dataset.

\begin{figure}
    \centering
    \begin{subfigure}[t]{0.3 \textwidth}
        \includegraphics[width=\linewidth]{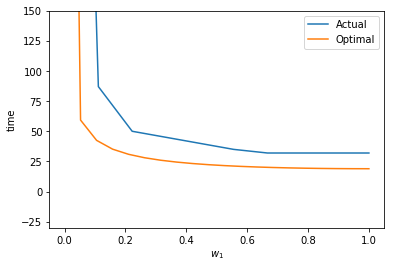}
        \caption{Time}
        \label{fig:timesteps}
    \end{subfigure}
    \begin{subfigure}[t]{0.3 \textwidth}
        \includegraphics[width=\linewidth]{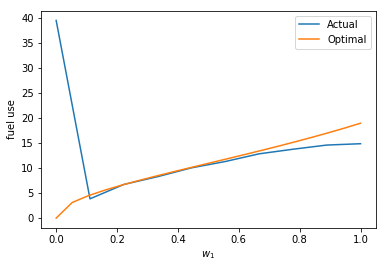}
        \caption{Fuel}
        \label{fig:fuel}
    \end{subfigure}
    \begin{subfigure}[t]{0.3 \textwidth}
        \includegraphics[width=\linewidth]{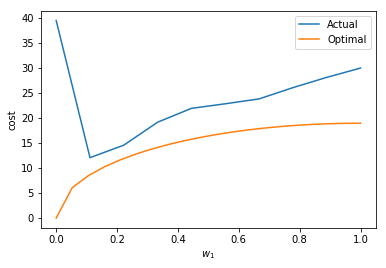}
        \caption{Cost (= negative reward)}
        \label{fig:reward}
    \end{subfigure}
    \caption{Results on 1D double-integrator as a function of $w_1$--the weight for prioritizing time to goal ($r_1$). Note that $w_2 = 1 - w_1$. The agent started from position $x = -90$ and needed to reach $x = 0$. The orange line shows the theoretical optimal solution and the blue line shows the solution achieved by the agent. For this experiment $k = 4$.
    \ref{fig:timesteps}) Total time to reach goal as a function of $w_1$ \ref{fig:fuel}) Total fuel used as a function of $w_1$ \ref{fig:reward}) Total cost incurred by agent as a function of $w_1$. }
    \label{fig:1D_results}
\end{figure}

\section{Environment}
We tested the algorithm on a double integrator environment.
The agent controls the acceleration of a particle moving along a line and tries to reach a specified goal position.
Every time step, the agent receives a two dimensional reward. The first component, $r_1$ encourages the agent to reach the goal quickly: $r_1 = -1$ until the agent reaches the goal, at which point $r_1 = 0$.
The second component encourages the agent to minimize fuel use: $r_2 = -|u|$ where $u \in [-1, 1]$ is the acceleration.

The optimal solution to this problem depends on the weighting of the two reward components $r = \left( \begin{smallmatrix} w_1 & w_2 \end{smallmatrix} \right) \left( \begin{smallmatrix} r_1 \\ r_2 \end{smallmatrix} \right) $.
If all the weight goes toward reaching the goal quickly, then the optimal solution takes the form of a bang-bang control system--the particle accelerates half the distance to the goal and decelerates the rest of the way.
This solution does not care about the increased fuel use that the extra speed requires.
If, however, more weight is put on minimizing fuel use, then the optimal solution is a bang-off-bang controller--the particle accelerates part of the way, coasts for a while, and then decelerates.

We also scaled this problem up to the N-dimensional case, in which the agent controls the acceleration along each axis of an N-dimensional space.
In order to make the N-dimensional case harder than N one dimensional cases, the total acceleration that the agent could apply was limited to $1$. In other words, $\sum_{i=0}^N |u_i| <= 1$.
The reward is a $2N$-dimensional vector, with two reward components for each axis.

\section{Results}

We used the DDPG algorithm \citep{DDPG} with policy and value networks each consisting of two hidden layers of $256$ neurons with ReLU activation functions, a sigmoid activation on the policy output, and a linear activation on the value network output.

\subsection{1D Results}
Figure \ref{fig:1D_results} compares the theoretical optimal results with those of the algorithm as a function of the weight on $r_1$.
As $w_1$ increases, the agent takes less time to reach the goal (Figure \ref{fig:timesteps}), which makes sense, since $w_1$ prioritizes a low travel time.
Note, that when $w_1$ is $0$, the time to reach the goal approaches infinity. This is because all the weight is on minimizing fuel use, so the best strategy is to not move at all.
The algorithm, however, outputs a small, constant acceleration and did not reach the goal, which is why fuel use is not $0$ (Figure \ref{fig:fuel}).

\begin{figure}
    \centering
    \begin{subfigure}[t]{0.3 \textwidth}
        \includegraphics[width=\linewidth]{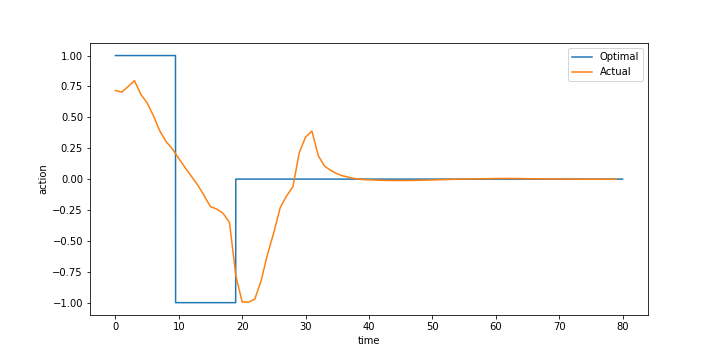}
        \caption{Optimal vs Learned Action}
        \label{fig:1Daction}
    \end{subfigure}
    \begin{subfigure}[t]{0.3 \textwidth}
        \includegraphics[width=\linewidth]{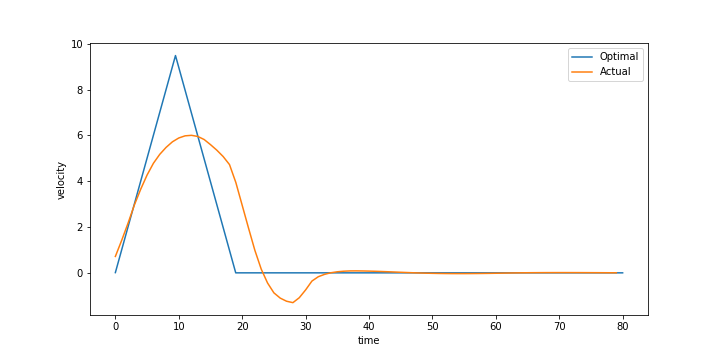}
        \caption{Optimal vs Learned Velocity}
        \label{fig:1Dvelocity}
    \end{subfigure}
    \begin{subfigure}[t]{0.3 \textwidth}
        \includegraphics[width=\linewidth]{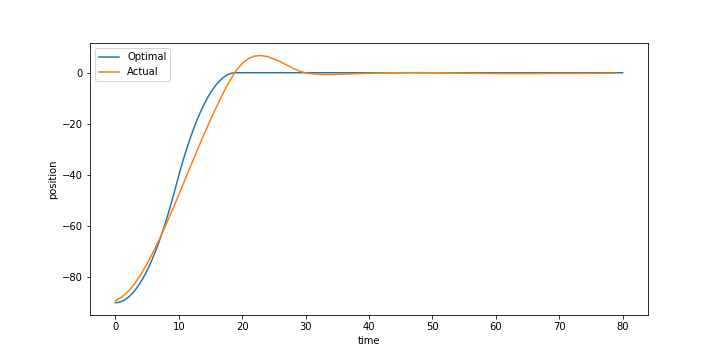}
        \caption{Optimal vs Learned Position}
        \label{fig:1Dposition}
    \end{subfigure}
    
    \caption{A sample run of the 1D double-integrator starting from position $x = -90$, with $\textbf{w} = (1, 0)^T$. Blue is the learned agent's action and orange is the optimal solution. \ref{fig:1Daction}) The action that the agent chose. \ref{fig:1Dvelocity}) The velocity of the agent over the course of the episode. \ref{fig:1Dposition}) The position of the agent over the course of the episode.}
    
    \label{fig:1D_sample_run}
\end{figure}

Except for the outlier at $w_1 = 0$, fuel use does increase as $w_1$ increases, as it should.
For smaller $w_1$, the optimal strategy would be to accelerate with maximum acceleration, then coast, and finally decelerate with maximum deceleration.
The algorithm instead learned to reduce fuel by minimizing its maximum acceleration.
The agent does not learn the optimal solution as can be seen in Figure \ref{fig:1D_sample_run}, but it does come close and it definitely learns an appropriate policy for each weight.

\subsection{2D Results}
In the 2D case, the agent controls the acceleration along two axes.
The reward, therefore, is a 4-dimensional vector: travel time along axis 1, fuel use of axis 1, travel time along axis 2, fuel use of axis 2.
The weights, $\textbf{w} = ( w_1, w_2, w_3, w_4)^T$ trade off between the different reward dimensions.
As the dimension of the weight vector increases, more samples should be necessary for the agent to learn to generalize because the volume of the weight space is bigger.
Despite, this the agent was able to learn the correct policy for each given weight.

\begin{figure}[t]
    \centering
    \begin{subfigure}[t]{0.35 \textwidth}
        \includegraphics[width=\linewidth]{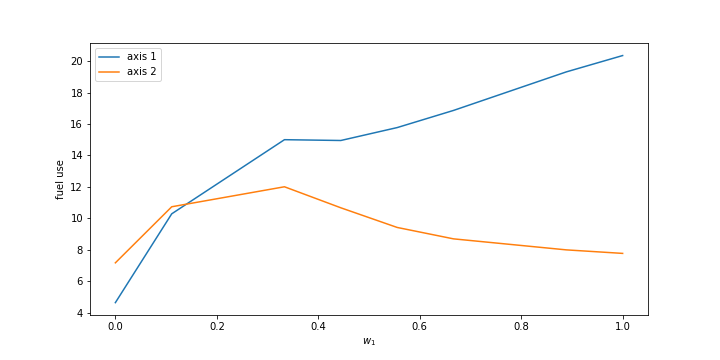}
        \caption{Fuel use: $w_1 = 0...1$ $w_2 = 1-w_1$}
        \label{fig:NDfuel12}
    \end{subfigure}
    \begin{subfigure}[t]{0.35 \textwidth}
        \includegraphics[width=\linewidth]{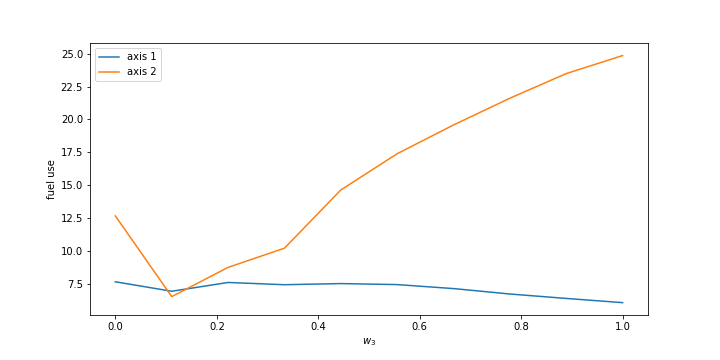}
        \caption{Fuel use: $w_3 = 0...1$ $w_4 = 1-w_3$}
        \label{fig:NDfuel34}
    \end{subfigure}
    \begin{subfigure}{0.35 \textwidth}
        \includegraphics[width=\linewidth]{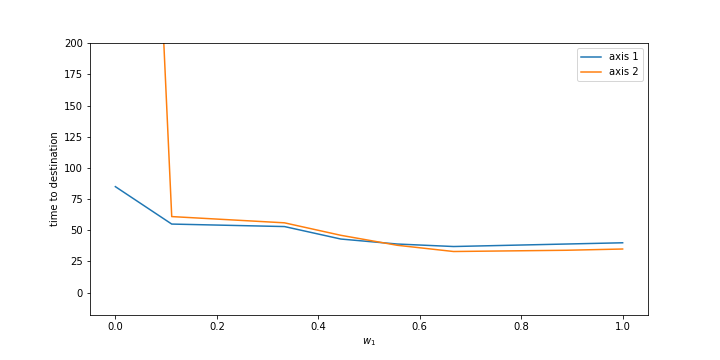}
        \caption{Time: $w_1 = 0...1$ $w_2 = 1 - w_1$}
        \label{fig:NDtime12}
    \end{subfigure}
     \begin{subfigure}{0.35 \linewidth}
        \includegraphics[width=\textwidth]{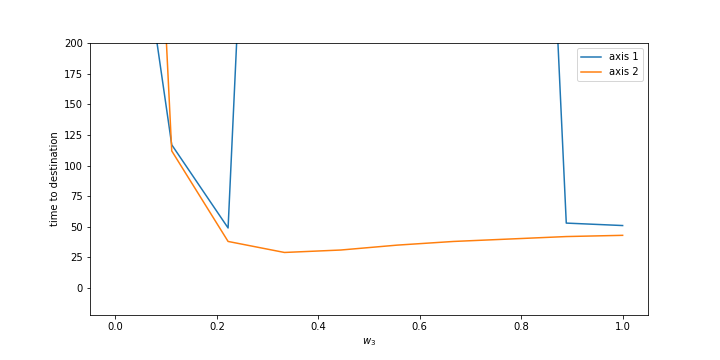}
        \caption{Time: $w_3 = 0...1$ $w_4 = 1 - w_3$}
        \label{fig:NDtime34}
    \end{subfigure}  
    
    \caption{Fuel use and travel time of agent in 2-D double-integrator. The agent started from position $x = -90$ and needed to reach $x = 0$. Figures \ref{fig:NDfuel12} and \ref{fig:NDtime12} show the total fuel use and travel time, respectively, of the agent along axis 1 (blue) and axis 2 (orange) as the axis 1 weights trade off between minimizing travel time ($\textbf{w} = (1, 0, 0, 0)^T$) and minimizing fuel use ($\textbf{w} = (0, 1, 0, 0)^T$). Figures \ref{fig:NDfuel34} and \ref{fig:NDtime34} show the same thing, but trading off between minimizing travel time ($\textbf{w} = (0, 0, 1, 0)^T$) and minimizing fuel use ($\textbf{w} = (0, 0, 0, 1)^T$) along axis 2.}
    
    \label{fig:ND_results}
\end{figure}

Figure \ref{fig:ND_results} shows results of the training as the weights trade off between minimizing fuel use and minimizing travel time along axis 1 (Figures \ref{fig:NDfuel12} and \ref{fig:NDtime12}) and axis 2 (Figures \ref{fig:NDfuel34} and \ref{fig:NDtime34}).
When the weight varies between minimizing fuel along axis 1 to minimizing time along axis 1, the fuel use of axis 1 increases and the time along axis 1 decreases.
The same is true when the weight varies on axis 2.
This behaviour makes sense, since more fuel is required to reach the destination more quickly.
Note that when the weights prioritize axis 1, the agent does not care about minimizing fuel use or time on axis 2. The same is true when prioritizing axis 2--in Figure \ref{fig:NDfuel34}, the fuel use along axis 2 increases, but along axis 1 remains flat, and in Figure \ref{fig:NDtime34} the time decreases along axis 2, but jumps up along axis 1.

\subsection{Augmentation Rate}
\begin{figure}[t]
    \centering
    \includegraphics[width = 0.7 \linewidth]{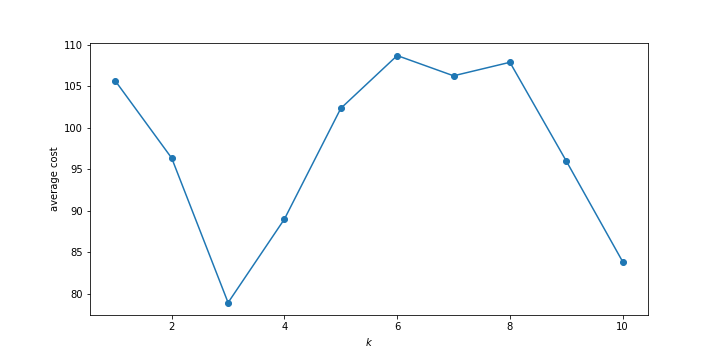}
    \caption{Augmentation rate, $k$, as a function of average cost}
    \label{fig:k_analysis}
\end{figure}
Our algorithm introduces one main hyperparameter, the augmentation rate $k$.
This parameter controls how many augmented experiences should be added to the replay dataset in relation to normal experiences.
For example, if $k=3$, then for every piece of data collected from the environment three augmented experiences are added.
For the 1D environment, we examined the cost received (negative reward) as a function of augmentation rate $k$, where the cost is averaged across all weights.
Figure \ref{fig:k_analysis} shows the results.
The cost is highly variable as a function of $k$, but there is a distinct minimum at $k=2$.
If $k$ is too low, then the algorithm would not have enough alternate experiences to generalize, but if $k$ is too high, then the algorithm is only learning from off-policy data, and learning would be slower.
\citeauthor{HER} \cite{HER}, however, report that in their experiments $k=4$ was optimal, which means that the optimal augmentation rate highly depends on the environment or the intended reward class that should be generalized.

\section{Conclusion}
We demonstrate that providing the reward function to the learning algorithm can be useful for teaching the agent to generalize across a distribution of rewards.
Specifically, we demonstrate that a neural network policy can learn to generalize across the weights of a linearly weighted, multi-objective reward function. 

Future work would extend these results to other classes of reward functions, such as learning to generalize across non-linearly weighted multi-objective rewards, or across different values of $\gamma$.



\newpage

\bibliography{references}

\end{document}